\documentclass[letterpaper, 10 pt, conference]{ieeeconf}
\IEEEoverridecommandlockouts                              
\usepackage{comment}
\usepackage{url}

\usepackage{multirow}
\usepackage{graphicx}
\usepackage{epsfig}
\usepackage{epic,eepic}
\usepackage{times}
\usepackage{mathptmx}
\usepackage{cite}
\usepackage{color}

\usepackage{times}

\definecolor{red}{rgb}{1,0,0}
\definecolor{green}{rgb}{0,1,0}
\definecolor{blue}{rgb}{0,0,1}
\definecolor{violet}{rgb}{1,0,1}
\definecolor{cyan}{cmyk}{1,0,0,0}
\definecolor{magenta}{cmyk}{0,1,0,0}
\definecolor{yellow}{cmyk}{0,0,1,0}

\definecolor{white}{rgb}{1,1,1}

\usepackage{arydshln}

\newcommand{\CO}[1]{}

\newcommand{\CommentOut}[1]{}

\newcommand{\noeditage}[1]{#1} \newcommand{\editage}[1]{}

\begin{document}

\newcommand{\FIG}[3]{
\begin{minipage}[b]{#1cm}
\begin{center}
\includegraphics[width=#1cm]{#2}\\
{\scriptsize #3}
\end{center}
\end{minipage}
}

\newcommand{\FIGU}[3]{
\begin{minipage}[b]{#1cm}
\begin{center}
\includegraphics[width=#1cm,angle=180]{#2}\\
{\scriptsize #3}
\end{center}
\end{minipage}
}

\newcommand{\FIGm}[3]{
\begin{minipage}[b]{#1cm}
\begin{center}
\includegraphics[width=#1cm]{#2}\\
{\scriptsize #3}
\end{center}
\end{minipage}
}

\newcommand{\FIGR}[3]{
\begin{minipage}[b]{#1cm}
\begin{center}
\includegraphics[angle=-90,width=#1cm]{#2}
\\
{\scriptsize #3}
\vspace*{1mm}
\end{center}
\end{minipage}
}

\newcommand{\FIGRpng}[5]{
\begin{minipage}[b]{#1cm}
\begin{center}
\includegraphics[bb=0 0 #4 #5, angle=-90,clip,width=#1cm]{#2}\vspace*{1mm}
\\
{\scriptsize #3}
\vspace*{1mm}
\end{center}
\end{minipage}
}

\newcommand{\FIGCpng}[5]{
\begin{minipage}[b]{#1cm}
\begin{center}
\includegraphics[bb=0 0 #4 #5, angle=90,clip,width=#1cm]{#2}\vspace*{1mm}
\\
{\scriptsize #3}
\vspace*{1mm}
\end{center}
\end{minipage}
}

\newcommand{\FIGpng}[5]{
\begin{minipage}[b]{#1cm}
\begin{center}
\includegraphics[bb=0 0 #4 #5, clip, width=#1cm]{#2}\vspace*{-1mm}\\
{\scriptsize #3}
\vspace*{1mm}
\end{center}
\end{minipage}
}

\newcommand{\FIGtpng}[5]{
\begin{minipage}[t]{#1cm}
\begin{center}
\includegraphics[bb=0 0 #4 #5, clip,width=#1cm]{#2}\vspace*{1mm}
\\
{\scriptsize #3}
\vspace*{1mm}
\end{center}
\end{minipage}
}

\newcommand{\FIGRt}[3]{
\begin{minipage}[t]{#1cm}
\begin{center}
\includegraphics[angle=-90,clip,width=#1cm]{#2}\vspace*{1mm}
\\
{\scriptsize #3}
\vspace*{1mm}
\end{center}
\end{minipage}
}

\newcommand{\FIGRm}[3]{
\begin{minipage}[b]{#1cm}
\begin{center}
\includegraphics[angle=-90,clip,width=#1cm]{#2}\vspace*{0mm}
\\
{\scriptsize #3}
\vspace*{1mm}
\end{center}
\end{minipage}
}

\newcommand{\FIGC}[5]{
\begin{minipage}[b]{#1cm}
\begin{center}
\includegraphics[width=#2cm,height=#3cm]{#4}~$\Longrightarrow$\vspace*{0mm}
\\
{\scriptsize #5}
\vspace*{8mm}
\end{center}
\end{minipage}
}

\newcommand{\FIGf}[3]{
\begin{minipage}[b]{#1cm}
\begin{center}
\fbox{\includegraphics[width=#1cm]{#2}}\vspace*{0.5mm}\\
{\scriptsize #3}
\end{center}
\end{minipage}
}

\newcommand{\tabA}{
\begin{table}[t]
\caption{Performacne results.}\label{tab:A}
  \centering
    \begin{tabular}{l|r}
    \hline
Planner & SPL \\\hline
Ours. & 0.413 \\
ALC & 0.404 \\
ON & 0.383 \\
RF & 0.359 \\\hline
\end{tabular}
\end{table}
}

\newcommand{\figA}{
\begin{figure}[t]
\hspace*{5mm}\FIG{8}{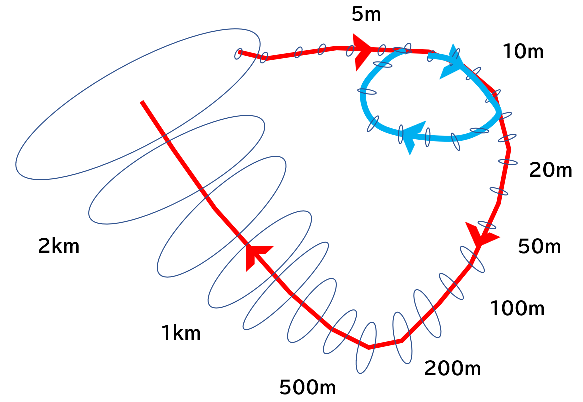}{}
\caption{Long-distance ALC (LD-ALC) problem.
The ellipse represents the error ellipse, and the numbers next to the curve are sample travel distances.
In contrast to the short-distance travel ALC scenario (blue line), the long-distance travel ALC scenario (red line) significantly impacts the map cumulative error reset caused by loop closing and presents a higher level of difficulty (LoD). This makes it both a practical and challenging problem, which is the focus of our interest.
}\label{fig:A}
\end{figure}
}

\newcommand{\figB}{
\begin{figure}[t]
\hspace*{5mm}\FIG{8}{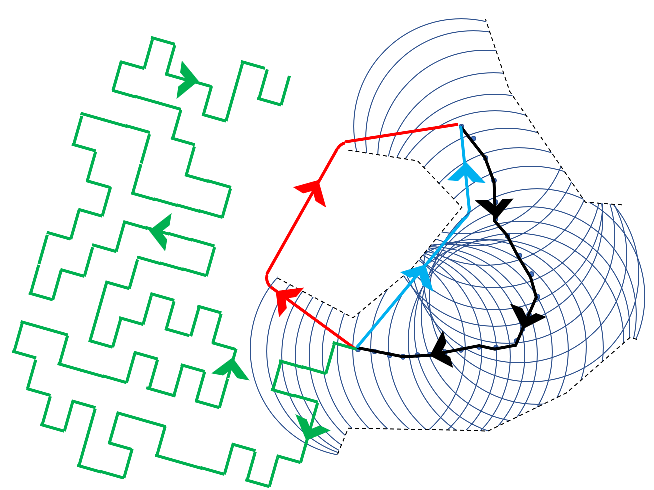}{}
\caption{ALC taxonomy.
The robot's viewpoint path is depicted as a black line with an arrow, and the area observed from each viewpoint without being occluded is illustrated as a circle.
(a)
Familiar place scenario (line with blue arrow): 
This represents an ALC scenario where familiar places are mainly traveled, and a closed-form solution exists.
(b)
Unfamiliar place scenario (line with red arrow): 
This represents an ALC scenario where unfamiliar places are mainly traveled, and it is highly non-trivial.
(c)
``Hard" or ``maze" scenario (line with green arrow): This represents a difficult ALC scenario in which the robot enters a maze-like domain, which requires a great deal of cost, and the map is contaminated by accumulated errors, and there is a high risk of ultimately failing to revisit PVPs.}\label{fig:B}
\end{figure}
}

\newcommand{\figC}{
\begin{figure}[t]
\hspace*{5mm}\FIG{8}{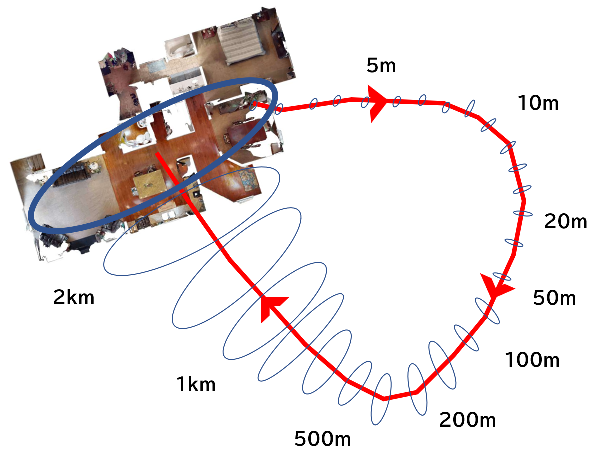}{}
\caption{Augmented ON workspace.
We generate the workspace for the ALC experiment by extending the standard workspace used in the object goal navigation (ON) literature. 
Instead of exhaustively generating a large virtual workspace that covers all viewpoints on a long-distance trajectory, we efficiently generate a workspace that covers only viewpoints near revisited points of interest in the ALC task. We then simulate this augmented ON workspace using Habitat-Sim, a standard ON simulator. The meaning of the error ellipse and travel distance sample values
follow Figure \ref{fig:A}.
}\label{fig:C}
\end{figure}
}

\newcommand{\figD}{
\begin{figure}[t]
\hspace*{5mm}\FIG{8}{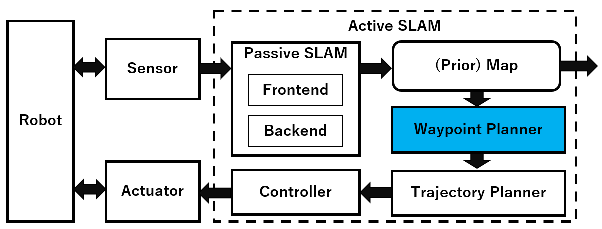}{\vspace*{3mm}a}
\hspace*{5mm}\FIG{8}{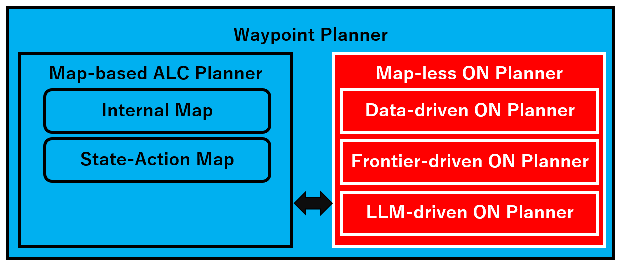}{\vspace*{3mm}b}
\caption{System overview. The ALC planner is formulated as an active SLAM system's waypoint planner module, which is responsible for planning waypoints (subgoals). (a) Role of the existing waypoint planner in the Active SLAM system. (b) Structure of the proposed extended waypoint planner. The map-less ON planner block, marked with a red box and white text, has been introduced.}\label{fig:D}
\end{figure}
}

\newcommand{\figE}{
\begin{figure}[t]
\hspace*{5mm}\FIG{8}{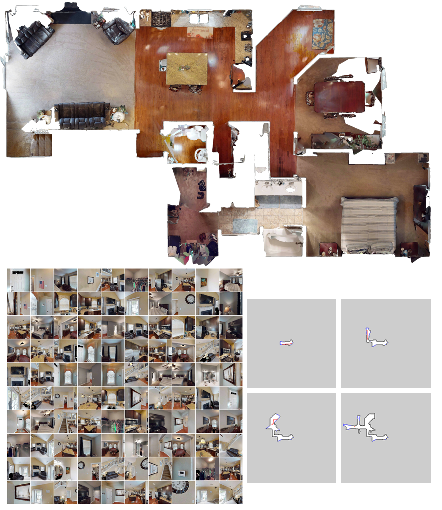}{}
\caption{Experimental Setup.
Top: Bird's-eye view of the workspace.
Bottom left: 100 target images. 
Bottom right: Illustration of the exploration and mapping process.
(Red curves: robot trajectories, Blue dots: subgoal candidates).
}\label{fig:E}
\end{figure}
}

\title{\LARGE \bf
ON as ALC: Active Loop Closing Object Goal Navigation 
}

\author{Daiki Iwata ~~~ Kanji Tanaka ~~~ Shoya Miyazaki ~~~ Kouki Terashima ~~~~\thanks{Our work has been supported in part by JSPS KAKENHI Grant-in-Aid for Scientific Research (C) 20K12008 and 23K11270.}\thanks{$*$%
D. Iwata, K. Tanaka, S. Miyazaki, K. Terashima are with Robotics Coarse, 
Department of Engineering, University of Fukui, Japan. 
{\tt\small{\{mf240050@g., tnkknj@, ha211418@g., mf240271@g.\}u-fukui.ac.jp}}}}

\noeditage{
\maketitle
}

\begin{abstract}
In simultaneous localization and mapping, active loop closing (ALC) is an active vision problem that aims to visually guide a robot to maximize the chances of revisiting previously visited points, thereby resetting the drift errors accumulated in the incrementally built map during travel. 
However, current mainstream navigation strategies that leverage such incomplete maps as workspace prior knowledge often fail in modern long-term autonomy long-distance travel scenarios where map accumulation errors become significant. To address these limitations of map-based navigation, this paper is the first to explore mapless navigation in the embodied AI field, in particular, to utilize object-goal navigation (commonly abbreviated as ON, ObjNav, or OGN) techniques that efficiently explore target objects without using such a prior map. Specifically, in this work, we start from an off-the-shelf mapless ON planner, extend it to utilize a prior map, and further show that the performance in long-distance ALC (LD-ALC) can be maximized by minimizing ``ALC loss" and ``ON loss". This study highlights a simple and effective approach, called ALC-ON (ALCON), to accelerate the progress of challenging long-distance ALC technology by leveraging the growing frontier-guided, data-driven, and LLM-guided ON technologies.
\end{abstract}

\section{Introduction}

In simultaneous localization and mapping (SLAM), active loop closing (ALC) is an active vision problem that aims to visually guide a robot to maximize the chances of revisiting previously visited points (PVPs), thereby resetting the drift errors accumulated in the incrementally built map during travel. Most existing works on SLAM assume a passive observer (robot) and does not address the issue of robot control (i.e., ALC). 
In certain application scenarios, such as small-scale environments where travel distances and accumulated map errors are not significant \cite{Neira2003}, or autonomous driving applications where robots inherently move along loop-like trajectories \cite{KITTI}, passive SLAM may even have the chance to encounter loop closures with sufficient frequency by chance.
However, in long-term autonomy and long-distance travel scenarios, where robots are expected to operate in environments such as homes, offices, hospitals, or disaster sites, which are characterized by high degrees-of-freedom action space and severe visual aliasing, the chances of encountering high-quality (100\% precision \cite{FABMAP}) loop closures significantly decrease. This makes the ALC problem increasingly critical (Fig. \ref{fig:A}), which this paper focuses on.

\noeditage{
\figA
}

In existing studies on ALC, the ALC action planner is typically triggered by an active SLAM system that incrementally builds a map (blue line in Fig. \ref{fig:A}) and relies on map-based navigation using the map as workspace prior knowledge. Examples of such map-based navigation include active self-localization \cite{ActiveLocInSurveyASLAM}, frontier-guided planning \cite{FrontierInSurveyASLAM}, information gain \cite{EntropyInSurveyASLAM}, and the Chinese postman problem (CPP) planner \cite{ChineeseInSurveyASLAM} (see reviews such as \cite{SurveyASLAM} for further details). However, in the long-distance travel scenarios, the accumulated map error becomes significant before the ALC action planner is triggered (Fig. \ref{fig:A}, red line), thereby considerably limiting the performance of map-based navigation. Additionally, several uncertainties, such as ``Which of the PVPs are revisit-able?", ``What is the minimum-cost traversable path to the PVP?", and ``To what extent has the environment changed?", complicate the ALC problem. Moreover, as map errors accumulate, the prediction of a target PVP's coordinate from the prior map becomes increasingly uncertain, necessitating broader exploration areas to locate the PVPs (Fig. \ref{fig:A}, ellipses). Consequently, the cost and risk of exploration rise dramatically. Furthermore, the impact of erroneous loop closures often results in catastrophic map collapse errors, and research on recovering from such failures in ALC is still in its infancy. Thus, ALC in long-distance travel scenarios remains highly unsolved.

To address these limitations of map-based navigation, this study explores the utilization of map-less navigation techniques, specifically object-goal navigation (commonly abbreviated as ON, ObjNav, or OGN), for long-distance ALC applications (LD-ALC). ON has recently emerged as one of the most active research areas in embodied AI \cite{OgnSurvey} and aims to explore target objects by leveraging sparse, data-driven, semantic, and commonsense reasoning cues without using prior maps in unseen and unfamiliar workspaces. For instance, \cite{ANS} introduces a photorealistic Habitat simulator to train data-driven reinforcement learning ON action planners. 
\cite{MiyzakiRef} extends frontier-based methods to efficiently search for targets based on semantic relationships with already detected other surrounding objects. 
\cite{CatShaped} leverages large language models (LLMs) such as ChatGPT as commonsense reasoning engines to boost ON capabilities. It is worth noting that such data-driven, frontier-guided, and LLM-guided ON approaches do not rely on prior maps and are not affected by accumulated map errors.

Instead of directly applying existing ON planners to ALC tasks, we argue that extending these planners to utilize the available prior maps as additional workspace knowledge in ALC tasks can significantly enhance ALC performance. 
In other words, this study makes a novel contribution, which we call ``ALCON" approach, which 
combines the best insights from the two worlds of map-based navigation (ALC) and mapless navigation (ON) \cite{TerashimaSII2024}.
First, unlike existing ON planners that build maps from scratch, we propose extending the ON planner concept to exploit prior maps. Furthermore, we introduce novel learning criteria, such as ``ALC loss" and ``ON loss" for supervised subgoal regression tasks, as well as ALC rewards and ON rewards for reinforcement learning-based subgoal planners. 
This framework excels at balancing the demands from the ON planner and ALC planner, with this balance managed by a single weight coefficient ($w_M$) that is automatically adjusted based on uncertainty estimates inherent in the prior maps.
Additionally, we augment the ON simulator (Habitat-Sim) to simulate challenging ALC scenarios, including LD-ALC tasks and recovery from catastrophic prior map failures. Comprehensive experiments using this augmented ON simulator validate the effectiveness of our proposed approach.
Our contributions are summarized as follows:
\begin{enumerate}
\item
We introduce a simple yet effective approach that extends off-the-shelf state-of-the-art object goal navigation (ON) planners to leverage prior maps.
\item
We propose novel ON/ALC loss/reward functions and strategically combine the strengths of both approaches based on the prior map's uncertainty estimation.
\item
We enhanced an ON simulator to build a long-distance ALC (LD-ALC) experimental environment and validated the proposed ALCON method.
\end{enumerate}

This study marks a first step toward a novel research direction that integrates map-based and mapless navigation for challenging long-distance ALC scenarios. 
It highlights a simple and effective approach to accelerate the development of ALC technologies by leveraging the multifaceted growth of ON technologies, including data-driven \cite{ANS}, frontier-guided \cite{MiyzakiRef}, and LLM-guided paradigms \cite{CatShaped}.

\noeditage{
\figD
}

\section{The ALC Problem}

Following the literature on active SLAM, the ALC planner is formulated as the ``waypoint planner" module in an active SLAM system. Figure \ref{fig:D} shows a typical active SLAM system for a vision-based wheeled mobile robot, where an onboard camera captures view images, a passive SLAM module reconstructs a map from view images, a waypoint planner globally plans sub-goals from the map, a path planner locally plans a trajectory to the sub-goals, and a controller navigates the robot along the trajectory. For detailed explanations, refer to literature such as \cite{SurveyASLAM}. 
Our focus, the waypoint planner, corresponds to the global planner in ON literature \cite{ANS}, 
which takes the map from an external passive SLAM module as input and determines sub-goals (waypoints).

\subsection{Prior Map}

In this paper, we refer to incrementally built maps provided by external passive SLAM modules (e.g., obstacle maps, cost maps, traversability maps) as prior maps. These are distinct from internally built maps such as those built from scratch by the ALC planner's internal representations (Section \ref{sec:OnPlanner}), implicit state-action maps, or score maps. As travel distance increases, the cumulative error in prior maps grows unbounded, and their utility diminishes accordingly. Generally, the level of difficulty of the ALC problem heavily depends on the type and accuracy of the prior map used. We believe that our approach can generalize to various types of maps, but in our experiments, we focus on a specific type of prior map, namely an obstacle grid map; see the experimental section (Section \ref{sec:exp}) for details.

\subsection{Taxonomy of ALC}

Broadly, the ALC problem can be further classified based on whether the path to the target PVP traverses observed known regions or not; the former is straightforward, whereas the latter is challenging.
The former involves navigating through observed areas to reach the target PVP (Figure \ref{fig:B}a), a basic ALC problem in active SLAM that has been frequently addressed (``homing subtask") and often has a closed-form solution \cite{ALC2024SOTA}.
The latter requires navigating through unobserved areas to reach the target PVP (Figure \ref{fig:B}b), which is inherently ill-posed and heavily relies on brute-force methods, such as vanilla frontier-guided planner approaches \cite{BruteForceFrontier}.

The worst-case scenario involves maps severely corrupted by drift errors, unknown regions structured like a maze, and high-cost exploration of previously unreachable PVPs that are blocked by dynamic obstacles, resulting in aimless retreat without reivisting a PVP, as illustrated in Figure \ref{fig:B}c. The likelihood of encountering such worst-case scenarios increases critically with travel distance and remains largely unresolved. Our primary interest lies in the latter case (Figures \ref{fig:B}b and \ref{fig:B}c).

\noeditage{
\figB
}

\section{Method}

This section provides a detailed explanation of how we extend the ON planner to the ALC planner.
First, the ON planner, a mapless navigation method, is introduced. Next, the ON planner is enhanced to utilize prior maps. Furthermore, novel ALC loss and ON loss are introduced to fine-tune the ON planner for both map-based and mapless navigation purposes. 
Unless otherwise specified, any "score map" mentioned in the following explanation of the action planner is represented as a 256-level grid cell with values in the range $[0, 1]$ and a spatial resolution of 0.1 m, and is implemented simply as a 480$\times$480 image with 256 grayscale levels.

\subsection{ON Formulation}\label{sec:OnPlanner}

In contrast to the ALC task which leverages prior maps as priors, the ON task typically assumes unfamiliar and unseen workspaces without depending on prior maps.

{\bf IIGN:} 
The goal of an ON task is to visually guide the robot to maximize the chances of reaching a target object (e.g., a mug, coffee maker, or kitchen) given as a query in the form of a view image.
The basic ON task requires exploration at the semantic level, whereas the LD-ALC scenario in this study is more closely related to the advanced instance image goal navigation (IIGN) task \cite{iign}, which involves exploring for a target object that matches the query object at the instance level.

{\bf Internal maps:} 
In contrast to external (active) SLAM systems that employ action planners, some form of internal (passive) SLAM module is often employed by action planners.
Note that knowledge transfer between the external SLAM sytem and an internal SLAM module can also be carried out via the ON action planner module (Fig. \ref{fig:D}b).
For example, the Active Neural SLAM \cite{ANS} system employs a neural SLAM as an internal SLAM module during ON tasks to build maps from scratch and use them for waypoint/trajectory planning. We refer to these maps managed by internal SLAM systems as ``internal maps"  (Fig. \ref{fig:D} ``internal map'') 
in the following, to distinguish them from the prior maps provided by external SLAM.

{\bf Multi-ON:} Viewing the ALC task as an ON problem, it is natural to treat the 
PVPs encountered along the robot's trajectory as the revisit point goal (RPG) candidates. Thus, our task is categorized as an advanced multi-ON task \cite{MultiON}, which deals with multiple targets rather than the basic single-target ON task. This advanced task has recently attracted increasing interest from researchers.

\subsection{Training Free Planner (TFP)}\label{sec:tfp}

Our planner is designed to take a score map computed by the state-of-the-art frontier-guided training-free planner (TFP) from \cite{MiyzakiRef} as input. The advantage of this TFP-guided method is its ability to compactly and effectively compress the robot's experience data into a single-channel score map representation. 

The score values of each grid cell in this score map are computed based on \cite{MiyzakiRef}.
These scores are determined by the similarity between the target image (e.g., coffee maker) and the input images at each viewpoint (e.g., detected kitchen) in the semantic embedding space of \cite{BERT}.
It is worth mentioning that in contrast to typical active SLAM that 
mainly utilizes the semantic features of the PVP object itself as a means to detect a PVP object, the state-of-the-art TFP method has the potential to use other surrounding objects that are semantically related to target objects as cues, potentially drastically improving the chances of detecting the target objects.
For example, when applied to the ALC task, the robot is expected to exhibit the following desirable ALC behavior:
(1) At the PVP viewpoint, the robot detects a coffeemaker alone, but not a kitchen. (2) At the RPG viewpoint, the robot fails to detect the coffeemaker due to its small size, but detects a kitchen because it is large and prominent. (3) Due to the strong semantic relationship between the kitchen and the coffeemaker, the robot strategically surveys the area around the kitchen with additional exploration cost. (4) Finally, the robot detects the coffeemaker near the kitchen, meaning the ALC was successfull.

TFP outputs clusters of semantic point clouds, so for our application, it is necessary to record them on the score map. For this purpose, we have adopted a simple scoring method where scores are assigned only to the centroids of each cluster, and these are recorded in the score map.

The TFP used in the experiment is the one implemented by us, based on \cite{MiyzakiRef}, with two key modifications described below.
(1) To apply it to our RGB camera setup instead of using an RGBD camera as in the above reference, we use MIDAS \cite{MIDAS} for (pseudo) depth image prediction.
(2) Unlike the aforementioned reference that uses three types of semantic embeddings, we simply used only a single embedding, BERT \cite{BERT}.

\subsection{Reinforcement Learning Planner (RLP)}

Building upon the TFP described above, a reinforcement learning planner (RLP) is trained as a global planner for subgoal determination, utilizing actor-critic reinforcement learning techniques within the Active Neural SLAM (ANS) framework \cite{ANS}, with several modifications outlined below:

(1)
Instead of using a specific eight-channel grid map as in ANS, a single-channel score grid map, such as the one described in Section \ref{sec:tfp}, is used as the input.

(2)
The trained RLP functions as a ``monitor" for the TFP heuristics planner \cite{MiyzakiRef}. 
This approach is motivated by the observation that both RLP and TFP have their own advantages and disadvantages; the RLP has the potential to constrain robot behavior based on long-horizon predictions, aiming for higher long-term rewards, but its spatial resolution of subgoal regression at each time step is relatively low, leading to dissatisfaction with the accuracy of planning when relying solely on the RLP; while the TFP provides much higher spatial resolution, though it is more myopic.
To combine the advantages of RLP and TFP, we propose a framework where the subgoal output from the TFP is monitored by the RLP, in the following procedure.
(1) After calculating the Euclidean distance between the highest-scoring subgoal predicted by TFP ($p_M$) and the subgoal regressed by RLP ($p_I$), if the distance exceeds a threshold of $G = 5$ [m], 
the TFP prediction is "adjusted" as follows:
\begin{equation}
p = p_M w_M + p_I (1-w_M),
\end{equation}
which has the effect of bringing the final subgoal closer to the RLP's prediction $p_I$. 
Otherwise, the TFP output is directly used as the final subgoal. 
The weighting parameter $w_M=0.7$ used by the framework is based on a naive approach and has not yet been optimized.

(3)
Additionally, instead of using imitation learning for the local planner to plan the trajectory to the subgoal as in ANS, a shortest path planner (Dijkstra's algorithm) is employed.

(4)
The advantage function in the actor-critic framework is defined as follows:
\begin{equation}
A = R + \gamma Q(s,a) - V(s),
\end{equation}
where $R$ is the immediate reward in reinforcement learning, and $Q(s,a)$ and $V(s)$ represent the state-action value and state value for state $s$ and action $a$, respectively. $A$ is the advantage function of the actor-critic model.
The actor model is a CNN that regresses 2D coordinates in the bird's-eye view coordinate system. Specifically, the CNN processes input images through two convolutional layers, followed by a fully connected layer, and ultimately outputs a 2D vector representing the subgoal location.
It is trained to regress subgoals with the highest score in a simulator (Section \ref{sec:sim}) where the robot receives a reward at each step of each episode. 
Common to $L_{ON}$ and $L_{ALC}$,
the Euclidean distance between the ground truth and predicted coordinates
is used as the loss. 
The critic model has a structure similar to the CNN architecture used by the actor model, but the architecture of the final layer is slightly modified to regress a one-dimensional state value. The loss is given as follows:
\begin{equation}
L = A^2 + k | V_{prediction} - V_{oracle} |
\end{equation}
where $k = 0.1$, and $V_{prediction}$ and $V_{oracle}$ are the predicted and ground-truth state-values, respectively.

(5)
Unless otherwise specified, the method for calculating the reward the robot receives when it reaches a certain coordinate is defined as the sum of scores within a disk of radius 2 m centered at that coordinate, which is then scaled by a factor of 0.001 to prevent excessively large values.

\subsection{Integrating ON and ALC}

Here, we consider integrating the prior maps used in the existing ALC task into the ON framework while accounting for the inherent map accumulation errors.

{\bf Loss/Reward:} 
The ALC loss $L_{ALC}$ is defined as the Euclidean distance $E_{ALC}$ between the target coordinate predicted by the ALC model and the corresponding ground-truth coordinate.
Similarly, the ON loss $L_{ON}$ is defined as the Euclidean distance $E_{ON}$ between the target coordinates predicted by the ON model and the ground truth coordinates. 
Given those loss functions, two RLPs $f_{ALC}$ and $f_{ON}$ are trained that use the ALC reward and the ON reward functions, where the ALC reward is the inverse of the ALC loss, the ON reward is the inverse of the ON reward, and the robot receives a reward at each step of a training episode.

Many workspaces are not necessarily rectangular, so when represented by a rectangular input grid map, the representation error often becomes significant. To minimize this negative impact, we introduce a method to reduce the area of invalid regions in the grid map that is input to the actor/critic CNN model. Specifically, instead of the naive approach of directly inputting the entire grid map into the CNN model, we calculate a bounding box that encloses only the mapped regions of the grid map. 
This bounding box is then used to crop a rectangular region from the grid map, which is further resized to $240 \times 240$ to match the input size of the actor/critic CNN model. Additionally, the score value is scaled from the range $[0, 255]$ to $[0, 1]$.
This adjustment was found to significantly improve performance.

The score map used to calculate the ON loss/reward is computed as follows. 
The score map generated based on \cite{MiyzakiRef} is notably noisy. To address this, only the top three positions in the score map are extracted, and score-weighted disks centered at these positions are used for scoring. 
We found that this method often suppresses noise and improves the reliability of the ON reward calculation.
Specifically, these three disks are with a fixed radius of $r'=2$ [m]. The score values within each disk were set to 
50, 150, and 255. Regions not covered by any disk were assigned a score of 0.

Given the uncertainty estimation $K$ for the RPG, the score map used to compute the ALC loss (ALC reward) is calculated as follows. An RPG is randomly placed within a circle of radius $K$, and a disk centered at this point is placed to generate the score map. Specifically, these disks are with a fixed radius of $r'=2$ [m]. The score values within each disk were set to 255. Regions not covered by any disk were assigned a score of 0.

It was observed that regression performance deteriorates if part of the disk extends beyond the boundaries of the score map. To address this issue, the score map is pre-expanded by a size $r = 3$ [m], which exceeds the disk radius.

If the regression model outputs coordinates outside the boundaries of the map, the regression result is deemed invalid. In cases where either the ALC regression model or the ON regression model produces invalid results, the regression result from the other model is adopted. 
The next-best-subgoal is determined by the weighted sum of the regression results from $f_{ALC}$ and $f_{ALC}$.
Specifically, the weight coefficient $w_{ALC}$ is calculated based on the uncertainty estimate $K$ available from
Then, the weighted average is computed as:
\begin{equation}
p = p_{ALC} \cdot w_{ALC} + p_{ON} \cdot (1 - w_{ALC}),
\end{equation}
and the frontier cell closest to $p$ in terms of Euclidean distance is determined as the next-best-subgoal.

\subsection{Efficiency Improvements in Training}\label{sec:sim}

{\bf DA:} Following the principles of Sim-to-Real, a specially designed data augmentation (DA) method is employed to simulate training episodes. 
Recall that the input modality of the action planner is the score map, represented as a single-channel image. Note that the distribution of such low-dimensional input modalities can be effectively represented with far fewer training samples compared to high-dimensional modalities. Specifically, in our simulator, a limited number of score maps are prepared, and augmented score maps are generated by repeatedly applying data augmentation to these maps. Similarly, based on the same concept, augmented obstacle maps are generated for the one-dimensional obstacle maps. 
These simulated individual obstacle maps and score maps are directly used to construct the dataset, as described below.

{\bf AR:} 
Since no dataset exists that is suitable for our LD-ALC scenario, we constructed a novel dataset (Fig. \ref{fig:C}). 
A single training/test episode depends on a triplet consisting of the workspace $O_{workspace}$, the target object $O_{goal}$, and the robot's initial position $O_{start}$. A naive way to sample such a triplet is to generate a ridiculously large workspace for each episode, capable of performing long-distance navigation.
Obviously, such an approach is costly and not scalable.
Therefore, 
we adopt an augmented reality (AR) approach, generating a workspace only for a ``small area" spatially close to the PVP or RPG, which is conceptually similar to the ``on-demand loading" technique commonly used in video games (e.g., RPG games).
In the context of LD-ALC, the term ``small area" referred to here is defined as the error ellipse area that encompasses the possible PVPs under the influence of drift errors caused by long-distance travel.
This definition is reasonable because data outside these error ellipse areas has no influence on the decision-making of either baseline methods or the proposed ALC planner. 
In the experiment, this error ellipse area was approximated by the area of a single Habitat-Sim workspace, as it simplifies the implementation.
Following this implementation, our LD-ALC task is formulated as an extended ON task, where the goal is to detect target PVPs under the constraint of 
the given prior map.
For each generated workspace sample, the robot's start viewpoint $O_{start}$ and target PVP viewpoint $O_{goal}$ are randomly sampled from the set of free cells in the workspace, under the constraint that there exists at least one traversable path between the two viewpoints, $O_{start}$ and $O_{goal}$.

\noeditage{
\figC
}

\section{Experiments}\label{sec:exp}

For the experiments, we utilized the workspace 00800-TEEsavR23oF from Habitat-Sim \cite{Habitat} by importing it into the simulator.

We prepared 100 PVP view image samples per workspace. Non-salient images such as walls, which are unsuitable for evaluating ALC performance, were manually excluded. Each test episode uses one workspace and one or more randomly selected PVP images. 
The robot is equipped with a front-facing camera and captures view images at each viewpoint on the trajectories and at the reached subgoals.
Following ALC literature, the predicted location of the target PVP viewpoint is assumed to be available on the map but subject to a maximum positional error of $K$ [m] due to map accumulation uncertainty. 
The magnitude of this uncertainty $K$ [m] is assumed to be available as an attribute of the prior map, for example, which can be easily calculated from the error covariance matrix 
that is an essential outcome of a classical Kalman filter-based (passive) SLAM.
The robot's initial position and the $K$ setting are randomized within the workspace for each test episode. The specific RPG image set consists of 100 images generated as described in our previous study \cite{ppniv2024}.
Figure \ref{fig:E} illustrates the experimental setup.

\figE

The implementation of a training simulator is currently not optimized and represents training score maps using a very simple score map. 
Specifically, score maps were generated by randomly placing three small disks, each with a fixed radius of $r' = 2\ \mathrm{m}$, under the condition that the center of each disk lies within a free cell.
The score values within each disk were set to 255, 150, and 50, in descending order.
Regions not covered by any disk were assigned a score of 
0.
A total of 3,000 training score maps were generated. For each score map, the centroid coordinate of the highest-scoring disk 
is calculated and used as the ground-truth subgoal location.
The gamma value was set to 0.9, and the learning rates for both the actor and critic models were 0.001.

Unless otherwise noted, the spatial resolution of all grid maps was set to 0.1 m.

The performance of a test ALC episode was evaluated using SPL (Success weighted by Path Length). SPL measures how efficiently the agent reaches a target PVP, reflecting both success rate and path efficiency.

The robot's field of view was a circular sector with a radius of 3.2 m and a field angle of 40 degrees. The exploration was considered successful and terminated when the target object came within 1.6 m of the robot. 
We follow a two-phase approach in \cite{MiyzakiRef}.
Specifically, common to all proposed methods, ablation methods, and baseline methods, a single training/test episode is divided into a first phase consisting of $S$=5 steps, followed by a subsequent second phase.
In the first phase, all methods use the random frontier planner (RF), while in the second phase, each method uses its own planner (RF/TFP/ON/Ours).

For performance comparison, we compared the proposed method (Ours) with two ablation methods (ON, ALC) and a random frontier (RF) baseline. The ablation methods ON and ALC were derived by removing the ALC model and the ON model, respectively, from the proposed method.

The SPL performances of the methods were shown in Table \ref{tab:A}.
The proposed method, assisted by the ON task, significantly outperformed all ablation methods and the baseline.
The performance validation of the full active SLAM tasks for long-distance travel and long-term autonomy is planned to be reported in a follow-up study.

\noeditage{
\tabA
}

\section{Applications}

In this study, we explore a unified framework for object navigation (ON) tasks that pursue two distinct goals: Active Loop Closing and Object Goal Navigation. This framework has recently been extended into a more general paradigm known as multi-objective ON.
Such a multi-objective ON framework is rapidly gaining importance, as it addresses two often conflicting demands frequently encountered in ON tasks:
(1) Visiting known landmark objects (e.g., a dish cabinet) is effective for increasing the likelihood of discovering target objects (e.g., a cat-shaped mug) that are semantically related to those landmarks;
(2) Exploring unknown areas to discover new landmark objects (e.g., a computer desk) is likewise effective for increasing the chance of finding the target object (e.g., a cat-shaped mug).
This simple example highlights the fundamental trade-off between exploration and exploitation of knowledge. More generally, however, the objectives of navigation may include a wide variety of goals, such as lost item retrieval, active change detection, search and rescue, exploration and mapping, teacher-student knowledge transfer, postal delivery, and so on.
To achieve multi-objective ON, simply combining training-free planners, reinforcement learning-based planners, and zero-shot planners is not sufficient. A more comprehensive approach to multi-objective optimization is required---an approach we plan to detail in a forthcoming publication.

\bibliography{reference} 

\begin{thebibliography}{10}

\bibitem{Neira2003}
Jos{\'{e}} Neira, Juan~D. Tard{\'{o}}s, and Jos{\'{e}}~A. Castellanos.
\newblock Linear time vehicle relocation in {SLAM}.
\newblock In {\em Proceedings of the 2003 {IEEE} International Conference on
  Robotics and Automation, {ICRA} 2003, September 14-19, 2003, Taipei, Taiwan},
  pages 427--433. {IEEE}, 2003.

\bibitem{KITTI}
Andreas Geiger, Philip Lenz, Christoph Stiller, and Raquel Urtasun.
\newblock Vision meets robotics: The kitti dataset.
\newblock {\em The International Journal of Robotics Research},
  32(11):1231--1237, 2013.

\bibitem{FABMAP}
Mark Cummins and Paul Newman.
\newblock Highly scalable appearance-only slam-fab-map 2.0.
\newblock In {\em Robotics: Science and systems}, volume~5, page~17. Seattle,
  USA, 2009.

\bibitem{ActiveLocInSurveyASLAM}
Dieter Fox, Wolfram Burgard, and Sebastian Thrun.
\newblock Active markov localization for mobile robots.
\newblock {\em Robotics and Autonomous Systems}, 25(3-4):195--207, 1998.

\bibitem{FrontierInSurveyASLAM}
Cyrill Stachniss, Dirk Hahnel, and Wolfram Burgard.
\newblock Exploration with active loop-closing for fastslam.
\newblock In {\em 2004 IEEE/RSJ International Conference on Intelligent Robots
  and Systems (IROS)(IEEE Cat. No. 04CH37566)}, volume~2, pages 1505--1510.
  IEEE, 2004.

\bibitem{EntropyInSurveyASLAM}
Nathaniel Fairfield and David Wettergreen.
\newblock Active slam and loop prediction with the segmented map using
  simplified models.
\newblock In {\em Field and Service Robotics: Results of the 7th International
  Conference}, pages 173--182. Springer, 2010.

\bibitem{ChineeseInSurveyASLAM}
Alberto Soragna, Marco Baldini, Dominik Joho, Rainer K{\"u}mmerle, and Giorgio
  Grisetti.
\newblock Active slam using connectivity graphs as priors.
\newblock In {\em 2019 IEEE/RSJ International Conference on Intelligent Robots
  and Systems (IROS)}, pages 340--346. IEEE, 2019.

\bibitem{SurveyASLAM}
Julio~A Placed, Jared Strader, Henry Carrillo, Nikolay Atanasov, Vadim
  Indelman, Luca Carlone, and Jos{\'e}~A Castellanos.
\newblock A survey on active simultaneous localization and mapping: State of
  the art and new frontiers.
\newblock {\em IEEE Transactions on Robotics}, 39(3):1686--1705, 2023.

\bibitem{OgnSurvey}
Jingwen Sun, Jing Wu, Ze~Ji, and Yu-Kun Lai.
\newblock A survey of object goal navigation.
\newblock {\em IEEE Transactions on Automation Science and Engineering}, 2024.

\bibitem{ANS}
Devendra~Singh Chaplot, Dhiraj Gandhi, Saurabh Gupta, Abhinav Gupta, and Ruslan
  Salakhutdinov.
\newblock Learning to explore using active neural slam.
\newblock In {\em International Conference on Learning Representations (ICLR)},
  2020.

\bibitem{MiyzakiRef}
Junting Chen, Guohao Li, Suryansh Kumar, Bernard Ghanem, and Fisher Yu.
\newblock {How To Not Train Your Dragon: Training-free Embodied Object Goal
  Navigation with Semantic Frontiers}.
\newblock In {\em Proceedings of Robotics: Science and Systems}, Daegu,
  Republic of Korea, July 2023.

\bibitem{CatShaped}
Vishnu~Sashank Dorbala, James~F Mullen~Jr, and Dinesh Manocha.
\newblock Can an embodied agent find your ``cat-shaped mug"? llm-based
  zero-shot object navigation.
\newblock {\em IEEE Robotics and Automation Letters}, 2023.

\bibitem{TerashimaSII2024}
Kouki Terashima, Kanji Tanaka, Ryogo Yamamoto, and Jonathan Tay~Yu Liang.
\newblock Active robot vision for distant object change detection: {A}
  lightweight training simulator inspired by multi-armed bandits.
\newblock {\em CoRR}, abs/2307.14105, 2023.

\bibitem{ALC2024SOTA}
Wei Gao, Zezhou Sun, Mingle Zhao, Chengzhong Xu, and Hui Kong.
\newblock Active loop closure for osm-guided robotic mapping in large-scale
  urban environment.
\newblock In {\em 2024 IEEE/RSJ International Conference on Intelligent Robots
  and Systems (IROS)}. IEEE, 2024.

\bibitem{BruteForceFrontier}
Brian Yamauchi.
\newblock A frontier-based approach for autonomous exploration.
\newblock In {\em Proceedings 1997 {IEEE} International Symposium on
  Computational Intelligence in Robotics and Automation CIRA'97 - Towards New
  Computational Principles for Robotics and Automation, July 10-11, 1997,
  Monterey, California, {USA}}, pages 146--151. {IEEE} Computer Society, 1997.

\bibitem{iign}
Guillaume Bono, Leonid Antsfeld, Boris Chidlovskii, Philippe Weinzaepfel, and
  Christian Wolf.
\newblock End-to-end (instance)-image goal navigation through correspondence as
  an emergent phenomenon.
\newblock In {\em The Twelfth International Conference on Learning
  Representations, {ICLR} 2024, Vienna, Austria, May 7-11, 2024}.
  OpenReview.net, 2024.

\bibitem{MultiON}
Haitao Zeng, Xinhang Song, and Shuqiang Jiang.
\newblock Multi-object navigation using potential target position policy
  function.
\newblock {\em IEEE Transactions on Image Processing}, 32:2608--2619, 2023.

\bibitem{BERT}
Jacob Devlin Ming-Wei~Chang Kenton and Lee~Kristina Toutanova.
\newblock Bert: Pre-training of deep bidirectional transformers for language
  understanding.
\newblock In {\em Proceedings of naacL-HLT}, volume~1. Minneapolis, Minnesota,
  2019.

\bibitem{MIDAS}
Reiner Birkl, Diana Wofk, and Matthias M{\"u}ller.
\newblock Midas v3. 1--a model zoo for robust monocular relative depth
  estimation.
\newblock {\em arXiv preprint arXiv:2307.14460}, 2023.

\bibitem{Habitat}
Manolis Savva, Abhishek Kadian, Oleksandr Maksymets, Yili Zhao, Erik Wijmans,
  Bhavana Jain, Julian Straub, Jia Liu, Vladlen Koltun, Jitendra Malik, et~al.
\newblock Habitat: A platform for embodied ai research.
\newblock In {\em Proceedings of the IEEE/CVF international conference on
  computer vision}, pages 9339--9347, 2019.

\bibitem{ppniv2024}
Kouki Terashima, Daiki Iwata, Kanji Tanaka, Shoya Miyazaki, and Jonathan Tay~Yu
  Liang.
\newblock Con: Continual object navigation via data-free inter-agent knowledge
  transfer in unseen and unfamiliar places, 2024.

\end{thebibliography}
\bibliographystyle{unsrt}

\end{document}